\documentclass[11pt, copyright, logo, secondlogo]{template/lark}
\usepackage[authoryear, sort&compress, round]{natbib}
\usepackage{bbm}
\usepackage{enumitem}
\setlist{nosep}
\usepackage{xspace}
\usepackage{graphicx}
\usepackage{tabularx}
\usepackage{booktabs}
\usepackage{subcaption}
\usepackage{xcolor}
\usepackage{hyperref}
\usepackage{verbatim}
\usepackage{float}
\usepackage{cleveref}
\usepackage{wrapfig}
\usepackage{makecell}
\usepackage{multirow}
\usepackage{fvextra} 
\usepackage[normalem]{ulem}
\usepackage{url}
\usepackage{tcolorbox}
\usepackage{listings}
\usepackage{fancyvrb}
\usepackage{xcolor}
\usepackage{tcolorbox}
\usepackage{pifont}
\tcbuselibrary{breakable,skins}
\tcbuselibrary{listings,skins,breakable}

\usepackage{amsmath}
\usepackage{amssymb}
\usepackage{algorithm}
\usepackage{algpseudocode}

\usepackage{amsmath,amsfonts,bm}

\def\eqref#1{equation~\ref{#1}}

\def\1{\bm{1}}

\DeclareMathAlphabet{\mathsfit}{\encodingdefault}{\sfdefault}{m}{sl}
\SetMathAlphabet{\mathsfit}{bold}{\encodingdefault}{\sfdefault}{bx}{n}

\title{From Trainee to Trainer: LLM-Designed Training Environment for RL with Multi-Agent Reasoning}

\author[1]{Chao Chen}
\author[2]{Chengzu Li$^{\dag}$}
\author[1]{Zhiwei Li}
\author[2]{Yinhong Liu$^{\dag}$}
\author[1,3]{Zhijiang Guo$^{\dag}$}

\affil[1]{LARK, HKUST (GZ)}
\affil[2]{University of Cambridge}
\affil[3]{HKUST}

\correspondingauthor{Chengzu Li (cl917@cam.ac.uk), Yinhong Liu (yl535@cam.ac.uk), Zhijiang Guo (zhijiangguo@hkust-gz.edu.cn),\textbf{}}
\githubpage{https://github.com/LARK-AI-Lab/Trainee-to-Trainer}
\modelweight{https://huggingface.co/LARK-Lab/Trainee2Trainer/}
\website{https://lark-ai-lab.github.io/trainee-to-trainer.github.io}

\begin{abstract}
Reinforcement learning pipelines for Large Language Model (LLM) training often rely on manually redesigned environments between stages, requiring practitioners to heuristically infer which configuration will best improve the current policy. To automate this process, we propose the
\textbf{LLM-as-Environment-Engineer} framework in which the current
policy model analyzes failure trajectories together with contextual
information and proposes modifications to the next-stage training
environment configuration. We also introduce MAPF-FrozenLake, a
controllable testbed whose generator exposes multi-dimensional
environment configurations, making it suitable for studying and
benchmarking environment redesign. On this testbed, we condition the
environment engineer on structured summaries of policy behavior,
failure cases, and environment statistics, from which it produces the
configuration for the next training stage. With Qwen3-4B as the
backbone, our framework achieves the strongest aggregate performance on
our benchmarks, outperforming larger proprietary LLMs
(e.g., GPT, Gemini) and fixed-environment training baselines. We further analyze which
forms of context are most effective, finding that successful
environment updates rely on failure evidence and preserve
configurations that already work. Interestingly, the current RL checkpoint serves as a
better environment engineer than the original base model, suggesting
that policy learning improves the model's ability to diagnose its
remaining weaknesses.
\end{abstract}

\begin{document}

\fancyhead[C]{\fontsize{9}{12}\selectfont\color{black}%
From Trainee to Trainer: LLM-Designed Training Environment for RL with Multi-Agent Reasoning}

\maketitle

\section{Introduction}
\label{sec:introduction}

In Reinforcement Learning (RL) pipelines for training Large Language Models (LLMs; \citealt{Bai2026KimiKV, openai2026gpt5}), the interaction environment determines which behaviors, failures, and exploration signals the policy will encounter during training. As a result, environment design strongly affects optimization efficiency, generalization, and the emergence of new capabilities \citep{zhang2025auto}. However, in current practice, improving the training environment is still largely a manual process \citep{xie2025logic}. Practitioners repeatedly inspect rollout logs and validation failures, form hypotheses about the model's current weaknesses, and \textit{manually} redesign the next-stage training environment. This workflow requires substantial expert effort and becomes increasingly difficult as RL training pipelines scale in complexity, highlighting the need for an automatically adaptive framework.

\begin{figure}[t]
\centering
\includegraphics[width=0.7\columnwidth]{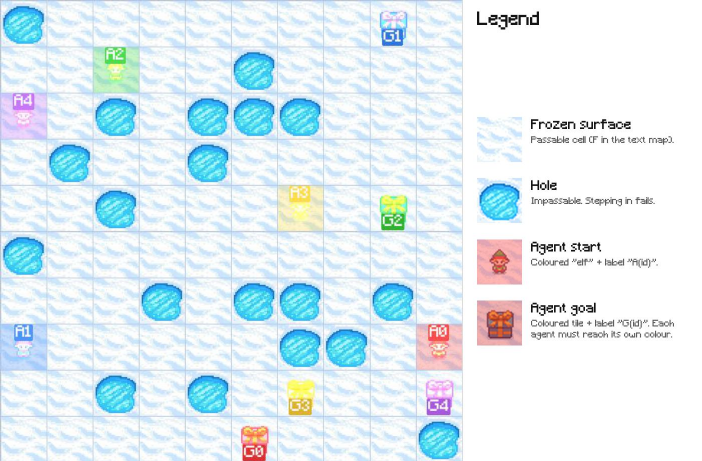}
\caption{A $5$-agent MAPF-FrozenLake instance on a $10{\times}10$ grid.
Agents A$0$/A$1$/A$2$/A$3$/A$4$ start at their colored cells and must reach
goals G$0$/G$1$/G$2$/G$3$/G$4$ without colliding with each other or stepping
into a hole (blue patches); a \emph{wait} action is available for
resolving conflicts.}
\label{fig:env_example}
\end{figure}

Recent work has automated related aspects of training adaptation. Curriculum learning~\citep{Azad2022CLUTRCL, bae2026online, xie2025logic} adjusts training difficulty over time, while self-play and multi-agent training frameworks~\citep{Fang2025SeRLSR, yuan2024self, shi2025mutual} adapt the training signal through interacting agents or opponents. More broadly, recent LLM-based data generation methods automatically construct synthetic training examples for downstream optimization~\citep{liang2025beyond}. However, these approaches mainly operate through training example selection, difficulty scheduling, or data synthesis within a fixed environment family. In contrast, we study a different problem: can the policy model itself proactively redesign the environment generator that defines its future RL training distribution?

To study this question, we propose a closed-loop framework for \textit{policy-conditioned environment redesign}, which we refer to as an \textbf{LLM-as-Environment-Engineer}. After each RL stage, the current policy model receives structured summaries of its training behavior, validation failures, and environment statistics, and proposes a new environment configuration for the next training stage. The model does not directly synthesize or select individual training examples. Instead, it modifies the parameters of an environment generator, thereby reshaping the future distribution from which training instances will be sampled. The objective of the environment engineer is therefore not to solve the task itself, but to propose training distributions that maximize future policy improvement.

This setting introduces several challenges. Validation signals are often sparse and highly dependent on environment configuration, so aggregated reward can hide distinct failure modes across different regions of the training distribution. Naively maximizing difficulty is also insufficient: overly difficult environments may collapse the learning signal, while overly easy environments can lead to premature saturation and weak exploration. Effective environment redesign therefore requires evidence-driven adaptation rather than shallow heuristics or monotonic difficulty scaling.

Studying these dynamics in open-ended embodied~\citep{Shridhar2020ALFWorldAT} or web environments~\citep{Zhou2023WebArenaAR} is difficult because the training distribution is only weakly controllable. Therefore, we design MAPF-FrozenLake, a controllable Multi-Agent Path Finding version of FrozenLake~\citep{Wu2025VSPDT} with grid-based environments, as a testbed for environment redesign. Figure~\ref{fig:env_example} shows a $5$-agent example on a $10{\times}10$ grid. Our goal is not to maximize environmental realism, but to isolate and study whether an LLM can make structured decisions about future training distributions under controlled conditions. Each instance is generated from a parameterized configuration that controls properties such as grid-size distribution, conflict density, and obstacle density. The environment also provides deterministic multi-dimensional evaluation signals, including path validity, optimality, and total trajectory length. These properties make MAPF-FrozenLake suitable for analyzing how environment redesign decisions affect downstream RL learning dynamics.

Combining the LLM-as-Environment-Engineer framework and MAPF-FrozenLake, we perform controlled investigations of environment redesign in RL-based LLM training. In particular, we study how different feedback signals influence redesign quality, whether RL training improves the model's ability to diagnose its own weaknesses, and which redesign behaviors are associated with successful downstream adaptation. Overall, our contributions are summarized as follows:

\begin{itemize}[leftmargin=*, itemsep=0pt, topsep=2pt, parsep=0pt]
    \item \textbf{Framework.} We propose a closed-loop framework where the policy LLM iteratively redesigns its own environment generator, and introduce MAPF-FrozenLake, a controllable testbed for studying this process.

    \item \textbf{Empirical Findings.} A 4B Qwen3 model under our framework outperforms both carefully-designed curricula and much larger proprietary LLMs (GPT, Gemini) as environment designers.

    \item \textbf{Mechanistic Insights.} RL training markedly improves the model's self-diagnostic ability: successful redesign is driven by evidence-based, failure-mode-targeted adaptation rather than naive difficulty scaling.
\end{itemize}
\section{Related Work}
\label{sec:related_work}

\noindent\textbf{Curriculum learning} improves training efficiency by controlling the
difficulty or ordering of training experience, and has been widely
applied across domains~\citep{bengio2009curriculum, graves2017automated}.
In reinforcement learning, many approaches assume a predefined set of
tasks and learn or specify a curriculum over them
\citep{li2023understanding, huang2022curriculum}. When such task sets
are unavailable, curricula are generated automatically by identifying
tasks near the current policy's capability boundary using signals such
as value estimates~\citep{zhang2020automatic, kim2023variational} or
episodic rewards~\citep{florensa2018automatic}. Recent LLM-based work
has adopted curriculum-style RL to improve reasoning and
generalization~\citep{bae2026online, zeng2503simplerlzoo}, while other
methods manually progress from easier to harder tasks after fixed
training stages~\citep{xie2025logic, team2025kimi}. Overall, existing
methods typically rely on auxiliary heuristics or fixed schedules to
select training tasks, limiting their ability to adapt the next-stage
environment directly to the model's observed failure patterns.


\noindent\textbf{Self-improvement} methods reduce reliance on external supervision by allowing models to generate their own training signals, such as tasks, responses, critiques, rewards, or opponents~\citep{gao2025survey}. Existing approaches mainly fall into two paradigms. Multi-model methods~\citep{shi2025mutual, huang2025r} train several models jointly so that one model provides challenges or feedback for another, while single-model methods~\citep{zhou2026self, yuan2024self, chen2025self, liang2025beyond} let a single model play multiple roles, including task proposal, solving, and evaluation. Across both paradigms, the goal is to create additional learning pressure without fully human-labeled data, typically within the task, response, or reward space. In contrast, our work focuses on the interaction environment itself: the model adapts the generator configuration that defines the situations encountered in the next RL stage.


\section{Method}
\label{sec:method}

We study whether an LLM can iteratively redesign its own future
training environment from detailed context. After each training round,
the model reviews the current state, decides how the next training
environment should change, and uses the resulting data for the next RL
stage. The central question is what context the model needs in order to
make these self-designed configurations useful.

\subsection{Task Formulation}
\label{sec:training_details}

\subsubsection{MAPF-FrozenLake Environment}
\label{sec:env}

MAPF-FrozenLake gives us a controlled but challenging setting for
studying environment design. Each instance places multiple agents on a
grid with holes; agents must reach their goals without collisions or
falling into holes, and may use \emph{wait} actions to resolve
conflicts. The environment can generate new instances from a
configuration, include map-size distribution, conflict ratio, and
hole density. Evaluation is also multi-dimensional, covering validity and total steps.

We build a environment generator on top of the Conflict-Based Search
algorithm. A generator configuration is denoted by $C$. For each map
size $m$, $C$ specifies a data ratio $C_{\text{data}}(m)$, a hole ratio
$C_{\text{hole}}(m)$, and a wait ratio $C_{\text{wait}}(m)$; these
values determine how many instances of each type are generated and how
difficult those instances are. To simulate a realistic cold start, the
initial training data is produced from a randomly sampled configuration. Let
$\mathcal{S} = \{3{\times}3, \ldots, 10{\times}10\}$
denotes the set of map sizes. For each map size $s$, a \emph{configuration}
\[
    C = \{\,(r_s,\, h_s,\, w_s)\,\}_{s \in \mathcal{S}}
\]
specifies, three components:
$r_s$, the \emph{data ratio} -- the share of training instances
sampled at size $s$, with $\sum_{s} r_s = 1$;
$h_s$, the \emph{hole ratio} -- the fraction of cells turned into
holes in maps of size $s$;
and $w_s$, the \emph{wait ratio} -- the fraction of generated
instances at size $s$ that require at least one wait action to
resolve agent conflicts. At round $R_n$, the configuration is denoted
$C_n$ and the generator emits the corresponding training set
$\mathcal{D}_n$.

\begin{figure*}[t]
\centering
\includegraphics[width=\linewidth]{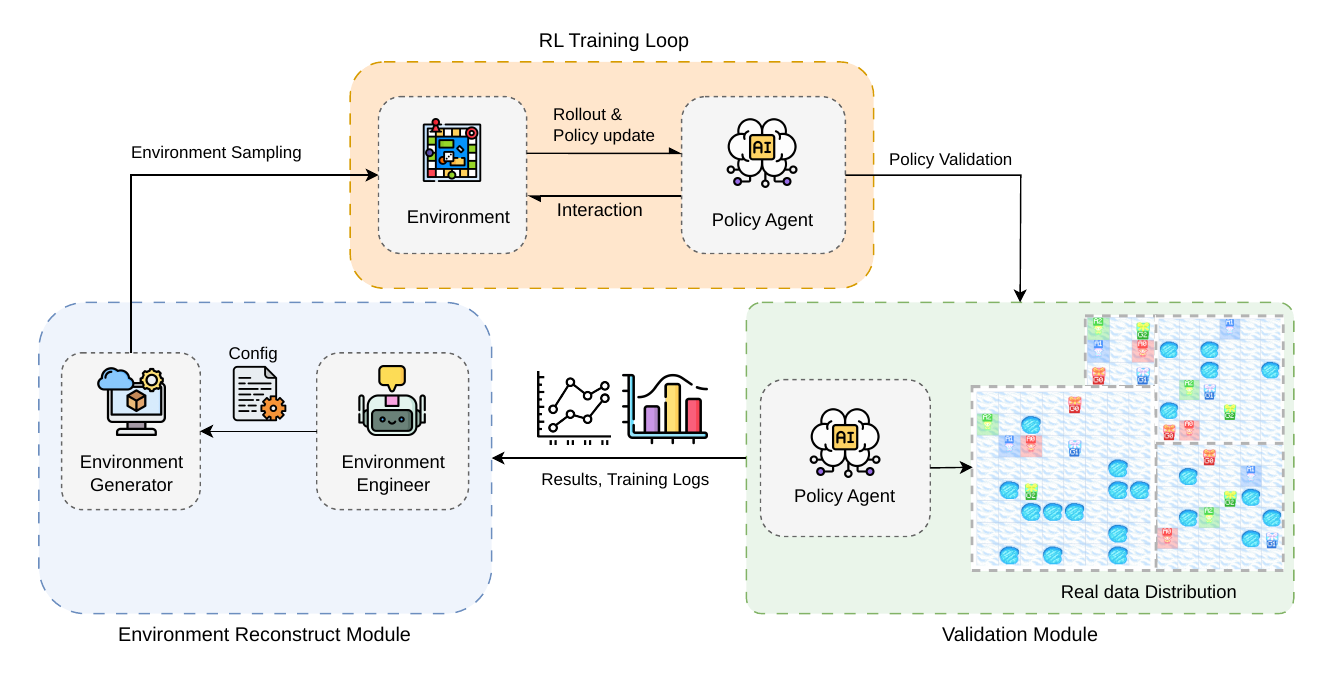}
\caption{Overview of environment-engineering framework. The model is
trained, validated, and then used as an environment engineer to
design the next-round training configuration.}
\label{fig:framework}
\end{figure*}

\subsubsection{RL Reward Design}
\label{sec:reward}

The total reward combines an accuracy term and a length term with
adaptive weights:
\begin{equation}
    R = w_{\text{acc}}\cdot R_{\text{acc}} + w_{\text{len}}\cdot R_{\text{len}}.
\end{equation}

\noindent\textbf{Accuracy reward.}
The response must pass eight strict validity checks; failing any one
sets $R_{\text{acc}}=0$. Details in
Appendix~\ref{app:validity_checks}.
If all eight pass, $R_{\text{acc}}$ is shaped by the cost gap to the
ground truth. Let $c = \mathit{model\_cost} - \mathit{gt\_cost}$ and
let $c_{\max} = 2\,\mathit{gt\_cost}$ be the saturation threshold; we
set
\begin{equation}
    R_{\text{acc}}(c) =
    \begin{cases}
        1.0, & c = 0, \\
        1.0 - 0.7\,\frac{c}{c_{\max}}, & c \in (0, c_{\max}], \\
        0.3, & c > c_{\max},
    \end{cases}
\end{equation}
which maps the cost-gap interval $[0, c_{\max}]$ linearly to the
reward interval $[1.0, 0.3]$ and saturates at $0.3$ beyond.
When $\mathit{gt\_cost}$ cannot be parsed we fall back to
$R_{\text{acc}}=0.5$.

\noindent\textbf{Length reward.}
We discourage verbose outputs. Let $\ell$ denote the response length
in tokens, and let $L_1 = 1500$ and $L_2 = 4096$ be the soft and hard
length thresholds; we set
\begin{equation}
    R_{\text{len}}(\ell) =
    \begin{cases}
        0, & \ell \le L_1, \\
        -\,\frac{\ell - L_1}{L_2 - L_1}, & L_1 < \ell < L_2, \\
        -1, & \ell \ge L_2,
    \end{cases}
\end{equation}
$R_{\text{len}}$ maps the length interval $[L_1, L_2]$ linearly to
reward interval $[0, -1]$ and saturates at $-1$ beyond.

\noindent\textbf{Adaptive weights.}
The two weights are scheduled to shift emphasis from brevity to
correctness as the model learns to produce concise outputs. Let $s$
denote the EMA of the short-response ratio (fraction of responses with
$\ell \le L_1$) across batches. We set
$(w_{\text{acc}}, w_{\text{len}}) = (0.5, 0.5)$ when $s < 0.5$,
linearly interpolate to $(0.8, 0.2)$ on $s \in [0.5, 0.9]$, and stay at
$(0.8, 0.2)$ once $s \ge 0.9$.

\subsection{Environment-Engineering Framework}
\label{sec:self_training_framework}

Unlike static curriculum design, our framework forms a closed feedback
loop in which the learner modifies the distribution of its future
training environments as its weaknesses change. As shown in
Figure~\ref{fig:framework}, each round consists of
\textsc{train}~$\to$~\textsc{eval}~$\to$~\textsc{design}. The model
used in \textsc{design} is the current learner checkpoint: it reads the
latest validation result, proposes the next configuration $C_{n+1}$,
and the generator produces the corresponding training data.

\begin{algorithm}[t]
\small
\caption{Environment-engineering train loop.}
\label{alg:env_engineering}
\begin{algorithmic}[1]
\Require base learner $M_0$, generator $G$, eval set $\mathcal{V}$,
initial config $C_0$, modules $\mathcal{M}\!\subseteq\!\{F,G,H,T\}$,
training meta $\mathcal{T}$, rounds $N$
\State history $\mathcal{B} \gets \emptyset$
\For{$n = 0$ \textbf{to} $N{-}1$}
    \State $\mathcal{D}_n \gets G(C_n)$ \Comment{generate env}
    \State $M_{n+1} \gets \textsc{GRPO}(M_n, \mathcal{D}_n; R)$
        
    \State $E_n \gets \textsc{Validate}(M_{n+1}, \mathcal{V})$
        
    \State $\mathcal{B} \gets \mathcal{B} \cup \{(C_n, E_n)\}$
    \State $\mathit{ctx}_n \gets
        \textsc{Compose}(\mathcal{M}; \mathcal{B}, \mathcal{T})$
    \State $\tilde{C}_{n+1} \gets M_{n+1}(\mathit{ctx}_n)$
        
    \State $C_{n+1} \gets \textsc{Project}(\tilde{C}_{n+1})$
        \Comment{enforce constraints}
\EndFor
\State \Return $M_N$, $\{C_n\}_{n=1}^{N}$
\end{algorithmic}
\end{algorithm}

The design context determines what evidence the environment engineer
can use when producing $C_{n+1}$. We study five modules. (1)
\textbf{Failure breakdown (F)} reports the latest validation outcome,
including aggregate valid and optimal rates and per-map-size counts for
parse errors, illegal moves, conflicts, hole collisions, out-of-bound
moves, and goal failures. (2) \textbf{Guideline (G)} provides
task-level design heuristics that are independent of a particular
round. (3) \textbf{History (H)} gives a short record of previous
$\{\textit{failure}, \textit{config}\}$ pairs; we compare versions with
and without the randomly sampled round-$0$ configuration. (4)
\textbf{Summary (S)} is a model-generated explanation of the current
configuration choice that is carried into later rounds. (5)
\textbf{Training details (T)} describes the RL objective already used
during training, including the reward design and adaptive-weight
schedule. It does not expose held-out instances or evaluation labels.
Because the environment engineer can only modify generator
configurations, 
this information supports training-aware environment design rather than
reward hacking.

\providecommand{\mvchk}{\textcolor{green!50!black}{\ding{51}}}
\providecommand{\mvxmk}{\textcolor{red!70!black}{\ding{55}}}

\begin{table*}[t]
\centering
\footnotesize
\setlength{\tabcolsep}{6pt}
\renewcommand{\arraystretch}{1.25}
\begin{tabular}{c c c c c c c}
\toprule
\textbf{Variant}
 & \shortstack{\textbf{Failure}\\\textbf{breakdown}}
 & \textbf{Guideline}
 & \shortstack{\textbf{History}\\\textbf{(w/ default config)}}
 & \shortstack{\textbf{History}\\\textbf{(w/o default config)}}
 & \textbf{Summary}
 & \shortstack{\textbf{Training}\\\textbf{details}} \\
\midrule
V1 & \mvchk & \mvxmk & \mvxmk & \mvxmk & \mvxmk & \mvxmk \\
V2 & \mvchk & \mvchk & \mvxmk & \mvxmk & \mvxmk & \mvxmk \\
V3 & \mvchk & \mvchk & \mvchk & \mvxmk & \mvxmk & \mvxmk \\
V4 & \mvchk & \mvchk & \mvxmk & \mvchk & \mvxmk & \mvxmk \\
V5 & \mvchk & \mvchk & \mvxmk & \mvchk & \mvchk & \mvxmk \\
V6 & \mvchk & \mvchk & \mvxmk & \mvchk & \mvxmk & \mvchk \\
\bottomrule
\end{tabular}
\caption{Modules included in each of the six context settings.}
\label{tab:variants}
\end{table*}

\begin{table*}[!t]
\centering
\footnotesize
\setlength{\tabcolsep}{3pt}
\renewcommand{\arraystretch}{1.05}
\resizebox{\textwidth}{!}{%
\begin{tabular}{l *{8}{cc} |cc}
\toprule
\multirow{2}{*}{\textbf{Model}}
 & \multicolumn{2}{c}{$3{\times}3$}
 & \multicolumn{2}{c}{$4{\times}4$}
 & \multicolumn{2}{c}{$5{\times}5$}
 & \multicolumn{2}{c}{$6{\times}6$}
 & \multicolumn{2}{c}{$7{\times}7$}
 & \multicolumn{2}{c}{$8{\times}8$}
 & \multicolumn{2}{c}{$9{\times}9$}
 & \multicolumn{2}{c}{$10{\times}10$}
 & \multicolumn{2}{|c}{Sum} \\
\cmidrule(lr){2-3}\cmidrule(lr){4-5}\cmidrule(lr){6-7}\cmidrule(lr){8-9}\cmidrule(lr){10-11}\cmidrule(lr){12-13}\cmidrule(lr){14-15}\cmidrule(lr){16-17}\cmidrule(lr){18-19}
 & acc.\ & opt.\ & acc.\ & opt.\ & acc.\ & opt.\ & acc.\ & opt.\
 & acc.\ & opt.\ & acc.\ & opt.\ & acc.\ & opt.\ & acc.\ & opt.\ & acc.\ & opt.\ \\
\midrule
GPT-5.4
 & 64.67 & 41.33 & 44.67 & 29.33 & 42.67 & 31.33 & 28.00 & 16.00
 & 20.67 & 12.67 & 26.67 & 15.33 & 20.00 & 11.31 & 12.67 & 7.33 & 32.50 & 20.58 \\
Grok-4.2
 & 36.00 & 25.33 & 50.00 & 37.33 & 29.33 & 18.67 & 35.33 & 21.33
 & 29.33 & 12.00 & 36.67 & 21.33 & 36.00 & 22.00 & 14.67 & 10.00 & 33.42 & 21.00 \\
Gemini-3.1-Pro
 & 45.33 & 33.33 & 32.67 & 23.33 & 35.33 & 20.00 & 24.00 & 12.00
 & 18.67 & 9.33 & 20.00 & 10.67 & 12.00 & 9.33 & 8.00 & 4.67 & 24.50 & 15.33 \\
Kimi-K2.5
 & 66.00 & 43.33 & 59.33 & 34.67 & 57.33 & 34.00 & 47.33 & 34.67
 & 38.00 & 22.67 & 41.33 & 24.00 & 36.67 & 24.67 & 23.33 & 16.00 & 46.17 & 29.25 \\
\midrule
Qwen3-4B (base)
 & 40.00 & 38.00 & 24.00 & 21.33 & 18.00 & 16.67 & 10.67 & 10.67
 & 8.00 & 8.00 & 5.33 & 4.67 & 10.00 & 10.00 & 2.67 & 2.67 & 14.83 & 14.00 \\
Qwen3-4B + GRPO (random)
 & 54.67 & 41.33 & 54.00 & 40.67 & 50.67 & 29.33 & 42.67 & 26.00
 & 38.67 & 21.33 & 30.67 & 18.00 & 32.67 & 18.67 & 19.33 & 13.33 & 40.42 & 26.08 \\
\textbf{Qwen3-4B + GRPO + Ours}
 & \textbf{68.67} & \textbf{48.00} & \textbf{64.67} & \textbf{41.33} & \textbf{62.67} & \textbf{35.33} & \textbf{52.67} & \textbf{37.33}
 & \textbf{46.00} & \textbf{26.00} & \textbf{42.67} & \textbf{24.00} & \textbf{44.00} & \textbf{25.33} & \textbf{32.00} & \textbf{16.00} & \textbf{51.67} & \textbf{31.67} \\
\bottomrule
\end{tabular}}
\captionof{table}{Main results on the \textbf{3-agent} evaluation set
across map sizes $3{\times}3$ to $10{\times}10$. \texttt{acc.} is the
valid rate (\%) and \texttt{opt.} is the optimal rate (\%); the
right-most \textbf{Sum} column reports the aggregate over all sizes.}
\label{tab:main_3agent}

\vspace{0.8em}
\resizebox{\textwidth}{!}{%
\begin{tabular}{l *{7}{cc} |cc}
\toprule
\multirow{2}{*}{\textbf{Model}}
 & \multicolumn{2}{c}{$4{\times}4$}
 & \multicolumn{2}{c}{$5{\times}5$}
 & \multicolumn{2}{c}{$6{\times}6$}
 & \multicolumn{2}{c}{$7{\times}7$}
 & \multicolumn{2}{c}{$8{\times}8$}
 & \multicolumn{2}{c}{$9{\times}9$}
 & \multicolumn{2}{c}{$10{\times}10$}
 & \multicolumn{2}{|c}{Sum} \\
\cmidrule(lr){2-3}\cmidrule(lr){4-5}\cmidrule(lr){6-7}\cmidrule(lr){8-9}\cmidrule(lr){10-11}\cmidrule(lr){12-13}\cmidrule(lr){14-15}\cmidrule(lr){16-17}
 & acc.\ & opt.\ & acc.\ & opt.\ & acc.\ & opt.\ & acc.\ & opt.\
 & acc.\ & opt.\ & acc.\ & opt.\ & acc.\ & opt.\ & acc.\ & opt.\ \\
\midrule
GPT-5.4
 & 35.33 & 22.00 & 24.00 & 16.00 & 16.00 & 10.00 & 14.00 & 9.33
 & 16.67 & 10.67 & 9.33 & 5.33 & 4.00 & 2.67 & 17.05 & 10.86 \\
Grok-4.2
 & 48.00 & \textbf{34.00} & 30.00 & 21.33 & 32.00 & 18.00 & 25.33 & 16.67
 & 5.33 & 3.33 & 8.67 & 6.67 & 14.67 & 7.33 & 23.43 & 15.33 \\
Gemini-3.1-Pro
 & 28.67 & 22.67 & 16.67 & 14.67 & 12.67 & 10.67 & 16.67 & 12.67
 & 8.67 & 5.33 & 4.67 & 2.00 & 3.33 & 2.67 & 12.95 & 10.10 \\
Kimi-K2.5
 & 44.67 & 28.67 & 35.33 & 20.67 & 28.67 & 16.67 & 28.00 & 22.67
 & 22.00 & 16.67 & 20.67 & 14.00 & 9.33 & 6.00 & 26.95 & 17.90 \\
\midrule
Qwen3-4B (base)
 & 10.67 & 9.33 & 4.67 & 4.00 & 2.67 & 2.67 & 4.00 & 3.33
 & 2.00 & 2.00 & 0.00 & 0.00 & 0.00 & 0.00 & 3.43 & 3.05 \\
Qwen3-4B + GRPO (random)
 & 42.67 & 27.33 & 33.33 & 23.33 & 31.33 & 18.67 & 26.67 & 15.33
 & 19.33 & 12.00 & 19.33 & 10.67 & 14.00 & 5.33 & 26.67 & 16.10 \\
\textbf{Qwen3-4B + GRPO + Ours}
 & \textbf{49.33} & 32.00 & \textbf{37.33} & \textbf{25.33} & \textbf{36.67} & \textbf{22.00} & \textbf{33.33} & \textbf{24.67}
 & \textbf{31.33} & \textbf{20.67} & \textbf{25.33} & \textbf{16.67} & \textbf{18.67} & \textbf{8.00} & \textbf{33.14} & \textbf{21.33} \\
\bottomrule
\end{tabular}}
\captionof{table}{Main results on the \textbf{4-agent} evaluation set
across map sizes $4{\times}4$ to $10{\times}10$. \texttt{acc.} is the
valid rate (\%) and \texttt{opt.} is the optimal rate (\%); the
right-most \textbf{Sum} column reports the aggregate over all sizes.}
\label{tab:main_4agent}

\vspace{0.8em}
\resizebox{\textwidth}{!}{%
\begin{tabular}{l *{6}{cc} |cc}
\toprule
\multirow{2}{*}{\textbf{Model}}
 & \multicolumn{2}{c}{$5{\times}5$}
 & \multicolumn{2}{c}{$6{\times}6$}
 & \multicolumn{2}{c}{$7{\times}7$}
 & \multicolumn{2}{c}{$8{\times}8$}
 & \multicolumn{2}{c}{$9{\times}9$}
 & \multicolumn{2}{c}{$10{\times}10$}
 & \multicolumn{2}{|c}{Sum} \\
\cmidrule(lr){2-3}\cmidrule(lr){4-5}\cmidrule(lr){6-7}\cmidrule(lr){8-9}\cmidrule(lr){10-11}\cmidrule(lr){12-13}\cmidrule(lr){14-15}
 & acc.\ & opt.\ & acc.\ & opt.\ & acc.\ & opt.\ & acc.\ & opt.\
 & acc.\ & opt.\ & acc.\ & opt.\ & acc.\ & opt.\ \\
\midrule
GPT-5.4
 & 17.33 & 14.00 & 10.00 & 6.00 & 10.00 & 4.00 & 6.00 & 4.67
 & 5.33 & 4.00 & 6.00 & 3.33 & 9.11 & 6.00 \\
Grok-4.2
 & 26.67 & 16.00 & 13.33 & 9.33 & 16.67 & 10.67 & 6.67 & 6.00
 & 7.33 & \textbf{5.33} & 6.67 & 5.33 & 12.89 & 8.78 \\
Gemini-3.1-Pro
 & 9.33 & 8.00 & 6.67 & 4.67 & 5.33 & 4.00 & 4.67 & 3.33
 & 2.00 & 1.33 & 0.67 & 0.00 & 4.78 & 3.56 \\
Kimi-K2.5
 & 23.33 & 17.33 & 18.00 & 12.00 & 16.00 & 7.33 & 14.00 & 8.67
 & 8.00 & 4.67 & 6.00 & 2.67 & 13.47 & 8.78 \\
\midrule
Qwen3-4B (base)
 & 2.67 & 2.00 & 2.67 & 2.67 & 0.67 & 0.00 & 2.00 & 2.00
 & 0.67 & 0.67 & 0.00 & 0.00 & 1.44 & 1.22 \\
Qwen3-4B + GRPO (random)
 & 24.00 & 16.67 & 21.33 & 14.00 & 16.00 & 10.00 & 13.33 & 8.00
 & 8.67 & 2.67 & 7.33 & 3.33 & 15.11 & 9.11 \\
\textbf{Qwen3-4B + GRPO + Ours}
 & \textbf{28.00} & \textbf{18.00} & \textbf{26.00} & \textbf{17.33} & \textbf{22.00} & \textbf{12.00} & \textbf{18.00} & \textbf{10.67}
 & \textbf{10.00} & 2.67 & \textbf{8.00} & \textbf{5.33} & \textbf{18.67} & \textbf{11.00} \\
\bottomrule
\end{tabular}}
\captionof{table}{Main results on the \textbf{5-agent} evaluation set
across map sizes $5{\times}5$ to $10{\times}10$. \texttt{acc.} is the
valid rate (\%) and \texttt{opt.} is the optimal rate (\%); the
right-most \textbf{Sum} column reports the aggregate over all sizes.}
\label{tab:main_5agent}

\end{table*}

\noindent\textbf{Context variants.}
We build up the context incrementally with 6 variants as summarized in Table~\ref{tab:variants}. 
V1--V3 add modules incrementally on top of the failure breakdown; V4 differs
from V3 only in that the round-$0$ default configuration is removed
from the history, so that the model does not treat the random default
as a recommended baseline; V5 and V6 then add the model-generated
summary (\textbf{S}) and the training-details module (\textbf{T}) on
top of V4. Implementation details of the prompts are provided in
Appendix~\ref{app:prompts}.

\noindent\textbf{Final framework.}
Based on the per-setting results, we adopt \textbf{V6} as the final
implementation of our framework. See Figure~\ref{fig:valid_rate} in
Appendix~\ref{app:valid_rate} for the detailed results. We analyze why this setting works
best in Section~\ref{sec:analysis_behavior}.
\section{Experiments and Analysis}
\label{sec:experiments}


\subsection{Setup}
\label{sec:setup}

\noindent\textbf{Data Construction and Evaluation.}
All training and validation data are produced by the generator
described in \S\ref{sec:env}. For each map size, the generator
configuration specifies the data ratio, the hole ratio, and the wait
ratio. Both the training set and the validation set consist
exclusively of $2$-agent instances, on grids ranging from
$3{\times}3$ to $10{\times}10$. We use a fixed total of $4000$
training instances in each round. Round $0$ uses a randomly sampled
configuration to produce both the initial training data and the
validation set; from round $1$ on, the configuration is emitted by
the environment engineer to adjust the training data, while the
validation set is held fixed across rounds.

\begin{figure*}[t]
\centering
\includegraphics[width=0.9\linewidth]{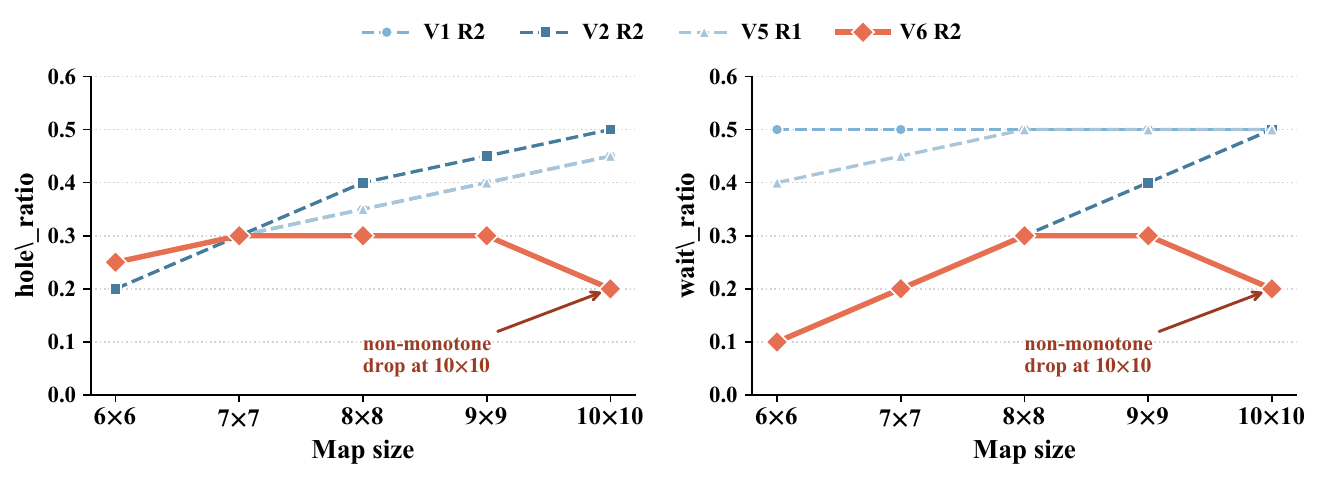}
\caption{\texttt{hole\_ratio} (left) and \texttt{wait\_ratio} (right)
on the five largest map sizes ($6{\times}6$--$10{\times}10$), where the
size-monotone vs.\ learner-aware contrast is sharpest. V1 R$2$,
V2 R$2$, and V5 R$1$ keep both ratios monotone in map size
(size-driven template), whereas V6 R$2$ plateaus at
$7{\times}7$--$9{\times}9$ and drops back at $10{\times}10$ on both
variables, where the failure breakdown indicates the largest maps are
no longer producing useful learning signal.}
\label{fig:template_vs_learner}
\end{figure*}

To probe generalization, we use a separate \emph{evaluation
benchmark} composed of $3$-, $4$- and $5$-agent instances. The
benchmark is divided into three subsets by wait ratio:
\texttt{wr\_025}, \texttt{wr\_050}, and \texttt{wr\_075},
corresponding to wait ratios $0.25$, $0.50$, and $0.75$. For each
map size within each subset, we use $50$ samples, evenly distributed
across the five hole ratios $\{0.1, 0.2, 0.3, 0.4, 0.5\}$. We use \emph{valid rate} and \emph{optimal rate}
as the evaluation metrics. The valid rate measures whether a response
parses to legal, conflict-free paths that reach the goals without
hitting holes or leaving the grid, matching the criteria used by the
accuracy reward in \S\ref{sec:reward}. The optimal rate measures
whether the model can find an optimal solution. These
two metrics together capture both correctness and solution efficiency.

\noindent\textbf{Implementation and Baselines.}

We use Qwen3-4B~\citep{Yang2025Qwen3TR} as the base model and train it with
verl, using GRPO with the adaptive-weight reward described
in \S\ref{sec:reward}. To balance training gains and computational efficiency,
we run three training rounds, each consisting of two epochs. We compare against four frontier LLMs: GPT-$5.4$,
Grok-$4.2$, Gemini-$3.1$-Pro and Kimi-K$2.5$, and two open-source
references that share our backbone: the untrained Qwen3-4B base
model and \emph{Qwen3-4B + GRPO (random)}, a Qwen3-4B trained with
the same GRPO procedure but on a single randomly sampled
Round-$0$ configuration. The latter isolates the contribution of the
training loop from the contribution of GRPO training itself. Detailed training hyperparameters are provided in
Appendix~\ref{app:training_details}.

\subsection{Main Results}
\label{sec:main_results}

\begin{figure}[t]
\centering
\includegraphics[width=0.87\columnwidth]{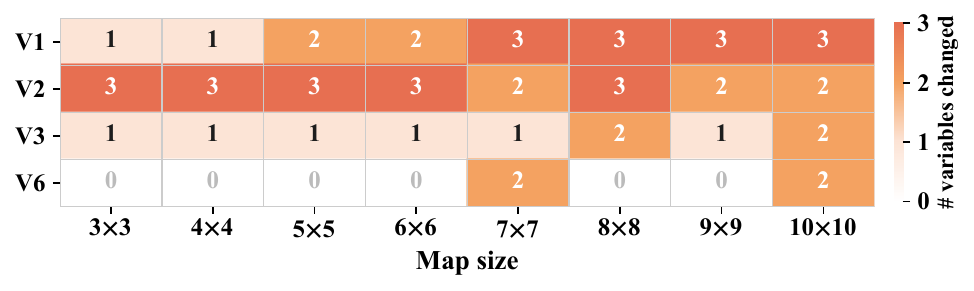}
\caption{Edit-granularity matrix. Each map size is controlled by three
variables: \texttt{data\_ratio}, \texttt{hole\_ratio}, and
\texttt{wait\_ratio}. Each cell reports how many of these three
variables the model changes between R$1$ and R$2$ for the given
variant and map size (threshold $|\Delta|{>}0.01$; possible values
$0$--$3$).}
\label{fig:edit_granularity}
\end{figure}

Tables~\ref{tab:main_3agent}, \ref{tab:main_4agent} and
\ref{tab:main_5agent} report the per-map-size valid rate
(\texttt{acc.}) and optimal rate (\texttt{opt.}) on the $3$-, $4$-
and $5$-agent evaluation sets. Across all three agent counts our
framework, \emph{Qwen3-4B + GRPO + Ours}, achieves the
highest aggregate Sum on both metrics, surpassing every closed- and
open-source baseline. Compared with the best commercial baseline (Kimi-K$2.5$) on each
agent count, across the $3$- to $5$-agent benchmarks our framework
improves the valid rate by $+5.20$ to $+6.19$ points and the optimal
rate by $+2.22$ to $+3.43$ points. Compared with
\emph{Qwen3-4B + GRPO (random)} -- which shares our backbone and
training procedure but uses a fixed configuration -- our
framework adds $+3.56$ to $+11.25$ points of valid rate and
$+1.89$ to $+5.59$ points of optimal rate across the three
benchmarks, showing that the training loop both raises the
model's overall capability and shapes the environment so that the
model more often produces the optimal plan, not merely a valid one.

\subsection{Behavioral Analysis}
\label{sec:analysis_behavior}

We analyze the environment engineer's reasoning traces across the
three training rounds (R$0$, R$1$, R$2$) and group the recurring
patterns into five behavioral dimensions; Table~\ref{tab:five_dim}
reports which dimensions each of the six context settings satisfies.

\begin{table}[t]
\centering
\footnotesize
\setlength{\tabcolsep}{4pt}
\renewcommand{\arraystretch}{1.15}
\begin{tabular}{c l c c c c c}
\toprule
\textbf{Rank} & \textbf{Setting}
 & \textbf{Salience}
 & \textbf{Granularity}
 & \textbf{Causal}
 & \textbf{Self-corr.}
 & \textbf{Task model} \\
\midrule
$1$ & \textbf{V6}
 & $\checkmark$ & $\checkmark$ & $\checkmark$ & $\checkmark$ & $\checkmark$ \\
$2$ & V4
 & $\checkmark$ & $\checkmark$ & $\times$ & $\checkmark$ & $\times$ \\
$3$ & V3
 & $\times$ & $\triangle$ & $\times$ & $\triangle$ & $\times$ \\
$4$ & V2
 & $\triangle$ & $\times$ & $\times$ & $\times$ & $\times$ \\
$5$ & V5
 & $\times\!\times$ & $\times$ & $\times$ & $\times$ & $\times$ \\
$6$ & V1
 & $\times\!\times$ & $\times$ & $\times$ & $\times$ & $\times$ \\
\bottomrule
\end{tabular}
\caption{Behavioral profile of the six context settings along the
five axes recovered from their reasoning traces.
$\checkmark$ = passes, $\triangle$ = partial,
$\times$ = fails, $\times\!\times$ = fails on multiple patterns of
the same axis. The left-most column gives the end-task ranking.}
\label{tab:five_dim}
\end{table}

\begin{figure*}[t]
\centering
\includegraphics[width=0.9\linewidth]{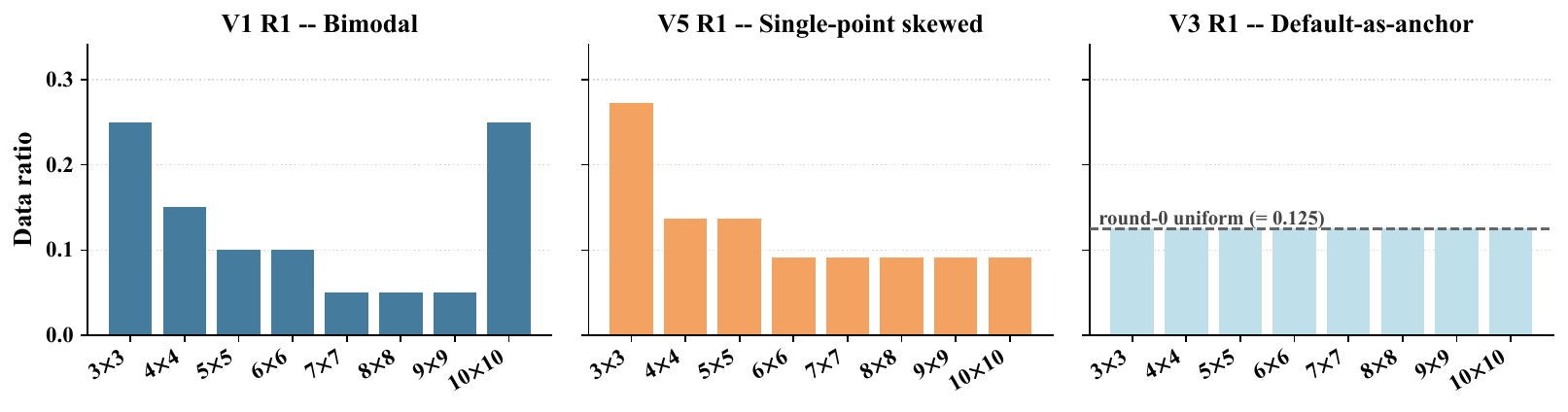}
\caption{Signal salience across context settings: each panel shows the
\texttt{data\_ratio} distribution emitted by a given (variant, round)
configuration. V1 R$1$ and V5 R$1$ spend the budget on summing to $1$;
V3 R$1$ stays close to the round-$0$ uniform default; V5 R$1{\to}$R$2$
lets its own self-summary override the raw failure breakdown.}
\label{fig:signal_salience}
\end{figure*}

\begin{figure*}[t]
\centering
\includegraphics[width=0.95\linewidth]{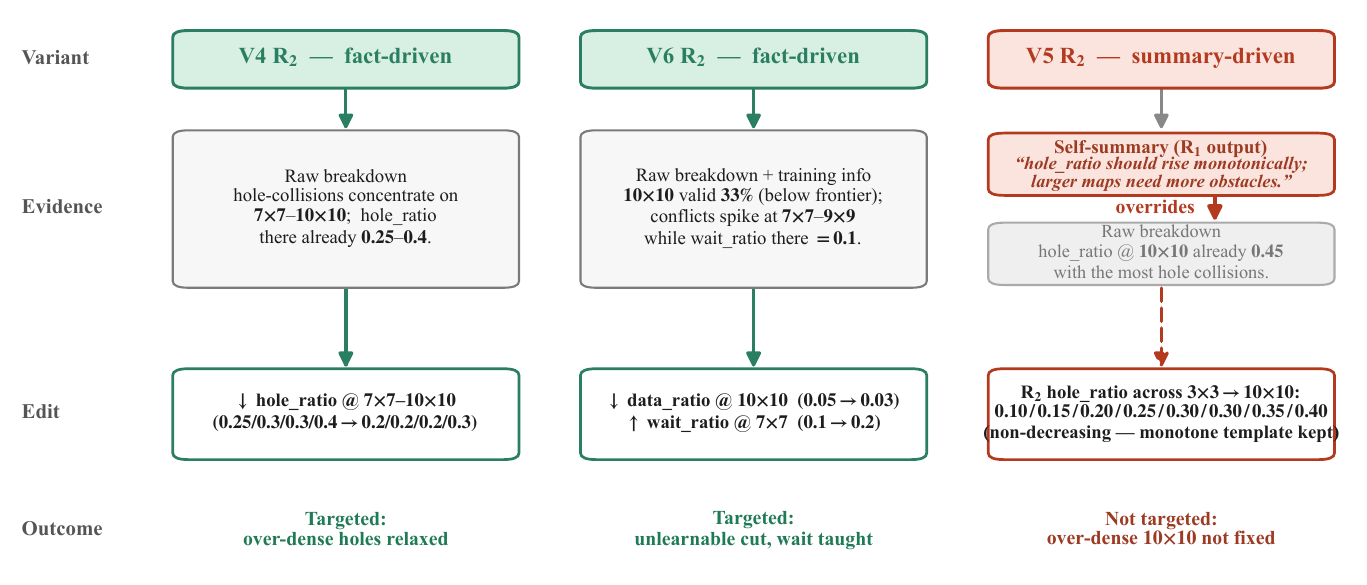}
\caption{Self-correction schematic. V4 R$2$ and V6 R$2$ edit the
configuration based on the raw failure breakdown, while V5 R$2$ lets
its own R$1$ self-summary override the breakdown and only nudges the
existing monotone template.}
\label{fig:self_correction}
\end{figure*}

\textbf{(1) Signal salience.} As shown in
Figure~\ref{fig:signal_salience}, the model is drawn to the most
locally certain cue inside each module: V1 R$1$ and V5 R$1$ spend
the budget on summing to $1$, V3 R$1$ stays close to the round-$0$
uniform default, and V5 R$1{\to}$R$2$ lets its own self-summary
override the raw failure breakdown. The shared mechanism is that the
most salient surface cue, rather than the most informative one,
drives the decision. We therefore remove the round-$0$ default and
the self-summary, and admit only modules whose dominant cue is
task-grounded.
\textbf{(2) Edit granularity.} From
Figure~\ref{fig:edit_granularity} we see that V1 and V2 rewrite
nearly every (size, knob) cell between rounds, while V3 and V6 leave
most sizes untouched and edit only the sizes the failure breakdown
points to. Full rewrites routinely degrade sizes that were already
healthy (V1 R$2$ raises the small-map hole ratios and small-map
valid rate drops), whereas selective edits do not. A good context
should therefore give the model the confidence \emph{not} to modify
what is already working.
\textbf{(3) Feature-based templates vs.\ learning-signal-based
decisions.} Figure~\ref{fig:template_vs_learner} shows that V1/V2/V5
keep both hole and wait ratios monotone in map size (the
\textit{``larger maps higher hole ratio''} template), whereas V6
plateaus at $7{\times}7$--$9{\times}9$ and drops back at $10{\times}10$
on both variables. The training-details module helps V6 reason about
which environments are likely to provide useful learning signal, rather
than simply mapping difficulty to surface map size. Effective contexts
therefore move the model from surface templates to updates that are
conditioned on the observed failure pattern.
\textbf{(4) Cross-round self-correction.}
Figure~\ref{fig:self_correction} shows that V4 R$2$ and V6 R$2$ edit
the configuration directly from the raw failure breakdown, whereas
V5 R$2$ lets its own R$1$ self-summary override the breakdown and
only nudges the existing monotone template. Self-correction therefore
requires evidence that is independent of the model's previous
narration. The context should provide facts as much as possible,
rather than only interpretations of those facts.
\textbf{(5) Task-grounded modeling.}
Table~\ref{tab:competence_frontier} shows that only V6 down-weights
the hardest map ($10{\times}10$ set to $0.05$ in R$1$) and
concentrates budget just below the competence frontier, then shifts
the frontier inward to $7{\times}7$ in R$2$. This behavior is
contingent on the training-details module: V1--V5 treat every size as
equally trainable and over-invest in the hardest size. Our framework
therefore avoids simply maximizing difficulty and instead uses the
observed failures to choose environments that remain useful for
training. Since the environment engineer can only change generator
configurations, not the reward function or evaluation set, this reflects
training-aware environment design rather than reward hacking.

\begin{table}[t]
    \centering
    \footnotesize
    \setlength{\tabcolsep}{5pt}
    \renewcommand{\arraystretch}{1.12}
    \begin{tabular}{c c c}
    \toprule
    \textbf{Map size}
     & \textbf{V6 R$1$ \texttt{data\_ratio}}
     & \textbf{V6 R$2$ \texttt{data\_ratio}} \\
    \midrule
    $3{\times}3$  & $0.05$  & $0.05$  \\
    $4{\times}4$  & $0.10$  & $0.10$  \\
    $5{\times}5$  & $0.15$  & $0.15$  \\
    $6{\times}6$  & $0.15$  & $0.15$  \\
    $7{\times}7$  & $0.15$  & $\mathbf{0.175}$ \\
    $8{\times}8$  & $\mathbf{0.175}$ & $0.175$ \\
    $9{\times}9$  & $\mathbf{0.175}$ & $0.175$ \\
    $10{\times}10$ & $\mathbf{0.05}$ & $\mathbf{0.03}$ \\
    \bottomrule
    \end{tabular}
    \caption{The shift of V6 \texttt{data\_ratio} from R$1$ to R$2$. The hardest map
    ($10{\times}10$) is down-weighted in both rounds; in R$2$ the budget
    shifts from the $8{\times}8$/$9{\times}9$ peak to $7{\times}7$, the
    new competence frontier.}
    \label{tab:competence_frontier}
    \end{table}

\begin{figure}[t]
\centering
\includegraphics[width=0.8\columnwidth]{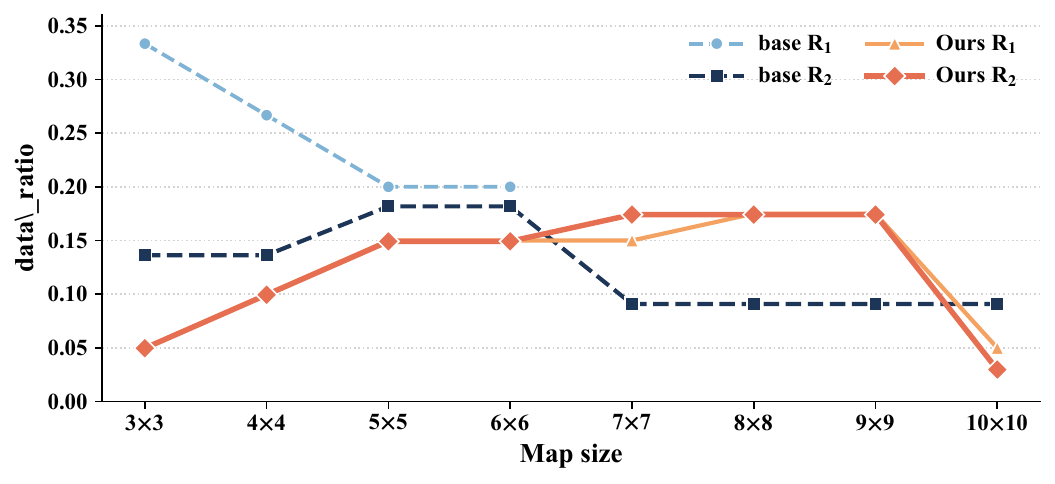}
\caption{\texttt{data\_ratio} produced by the base model
and by our current RL checkpoint when each plays the environment
engineer, across rounds R$1$ and R$2$.}
\label{fig:engineer_config}
\end{figure}

\begin{table}[t]
\centering
\footnotesize
\setlength{\tabcolsep}{4pt}
\renewcommand{\arraystretch}{1.15}
\begin{tabular}{l cc cc cc}
\toprule
\multirow{2}{*}{\textbf{Setting}}
 & \multicolumn{2}{c}{\textbf{3-agent}}
 & \multicolumn{2}{c}{\textbf{4-agent}}
 & \multicolumn{2}{c}{\textbf{5-agent}} \\
\cmidrule(lr){2-3}\cmidrule(lr){4-5}\cmidrule(lr){6-7}
 & acc.\ & opt.\ & acc.\ & opt.\ & acc.\ & opt.\ \\
\midrule
Full RL details         & $38.83$ & $26.50$ & $24.76$ & $18.19$ & $14.22$ & $10.11$ \\
\textbf{Bookkeeping only (Ours)}
                        & $\mathbf{51.67}$ & $\mathbf{31.67}$
                        & $\mathbf{33.14}$ & $\mathbf{21.33}$
                        & $\mathbf{18.67}$ & $\mathbf{11.00}$ \\
\bottomrule
\end{tabular}
\caption{Comparison between \emph{Full RL details} and
\emph{Bookkeeping only} on the $3$-, $4$- and $5$-agent benchmarks.}
\label{tab:ablation_training_details}
\vspace{-2mm}
\end{table}

\begin{table}[t]
\centering
\footnotesize
\setlength{\tabcolsep}{4pt}
\renewcommand{\arraystretch}{1.15}
\begin{tabular}{l cc cc cc}
\toprule
\multirow{2}{*}{\textbf{Engineer}}
 & \multicolumn{2}{c}{\textbf{3-agent}}
 & \multicolumn{2}{c}{\textbf{4-agent}}
 & \multicolumn{2}{c}{\textbf{5-agent}} \\
\cmidrule(lr){2-3}\cmidrule(lr){4-5}\cmidrule(lr){6-7}
 & acc.\ & opt.\ & acc.\ & opt.\ & acc.\ & opt.\ \\
\midrule
Untrained base                  & $45.21$ & $30.00$ & $27.62$ & $19.62$ & $16.00$ & $10.89$ \\
\textbf{Current checkpoint (Ours)}
                                & $\mathbf{51.67}$ & $\mathbf{31.67}$
                                & $\mathbf{33.14}$ & $\mathbf{21.33}$
                                & $\mathbf{18.67}$ & $\mathbf{11.00}$ \\
\bottomrule
\end{tabular}
\caption{Comparison between the untrained base model and the
current RL checkpoint playing the environment-engineer role,
on the $3$, $4$, $5$-agent benchmarks.}
\label{tab:ablation_engineer}
\end{table}

\subsection{Ablation Study}
\label{sec:ablation}

\noindent\textbf{What kind of training details matter (T).}
We ablate the training-details module to test whether V6 needs the full
RL training details or only basic loop bookkeeping. Both settings include
the current round index, epochs per round, and total epochs; only the
full setting additionally includes the GRPO algorithm, adaptive-weight
reward, and key hyperparameters. As shown in Table~\ref{tab:ablation_training_details}, the
bookkeeping-only setting outperforms the full setting on every
agent count, in both valid rate and optimal rate. This suggests that
the model mainly needs to know where it is in the training loop,
while detailed RL parameters can distract it from the current failure
evidence. In our framework, training details are useful when they
provide stage awareness, not when they overload the context with
optimization details.

\noindent\textbf{Who plays the environment engineer.}
We compare two choices for playingthe
environment-engineer role: the current RL checkpoint of the learner,
versus the untrained base model. As reported in
Table~\ref{tab:ablation_engineer}, with every other component of the
framework held fixed, the current checkpoint outperforms the base
model on all $3$-, $4$- and $5$-agent benchmarks in both valid rate
and optimal rate. Figure~\ref{fig:engineer_config} shows the
underlying \texttt{data\_ratio} allocations: in R$1$ the base
engineer abandons $7{\times}7$--$10{\times}10$ and pours the entire
budget into $3{\times}3$--$6{\times}6$, and in R$2$ -- even after the
larger sizes are added back -- it still gives each of them only
$\sim$$9\%$ of the budget, while our checkpoint maintains a
frontier-aware allocation across all eight sizes throughout both
rounds. We interpret this as a form of
\emph{self-aware learning}: policy learning sharpens the model's ability to diagnose its own
remaining weaknesses, so the trained checkpoint can target the next
stage's data more precisely at those gaps than the untrained base
model at the start.

\section{Conclusion}
\label{sec:conclusion}
We introduced a closed-loop framework where an LLM acts as an environment engineer, proactively redesigning the training configuration for its own RL learning. We developed MAPF-FrozenLake as a controllable testbed and showed that a 4B policy model, guided by structured feedback, can iteratively propose environment configurations that consistently outperform larger proprietary LLMs on Multi-Agent Path Finding tasks. Our mechanistic analysis further reveals that RL training enhances the model’s ability to diagnose its own weaknesses, and that successful redesign depends on evidence-driven adaptation rather than naive difficulty maximization. These findings lay a foundation for studying self-improving learning systems via policy-conditioned environment engineering.





\clearpage

\bibliography{custom}
\bibliographystyle{abbrvnat}

\clearpage

\appendix
\section{Validity checks for the accuracy reward}
\label{app:validity_checks}

This section lists the full set of validity checks that determine
$R_{\text{acc}}$. A response is counted as a success only if it
passes all eight of the following checks; failing any one sets
$R_{\text{acc}}=0$:
\begin{enumerate}
    \item \textbf{Parsable} -- paths can be extracted from the response;
    \item \textbf{Legal-move} -- each step has Manhattan distance $\le 1$;
    \item \textbf{Conflict-free} -- no vertex or edge conflict;
    \item \textbf{Goal-reached} -- every agent ends at its goal;
    \item \textbf{Start-correct} -- every agent starts at its start;
    \item \textbf{Agent-count} -- the parsed agent count matches the problem;
    \item \textbf{Hole-free} -- no path crosses a hole;
    \item \textbf{In-bounds} -- no path leaves the grid.
\end{enumerate}

\section{Per-setting valid rate of the six context variants}
\label{app:valid_rate}

\begin{figure*}[t]
\centering
\includegraphics[width=\linewidth]{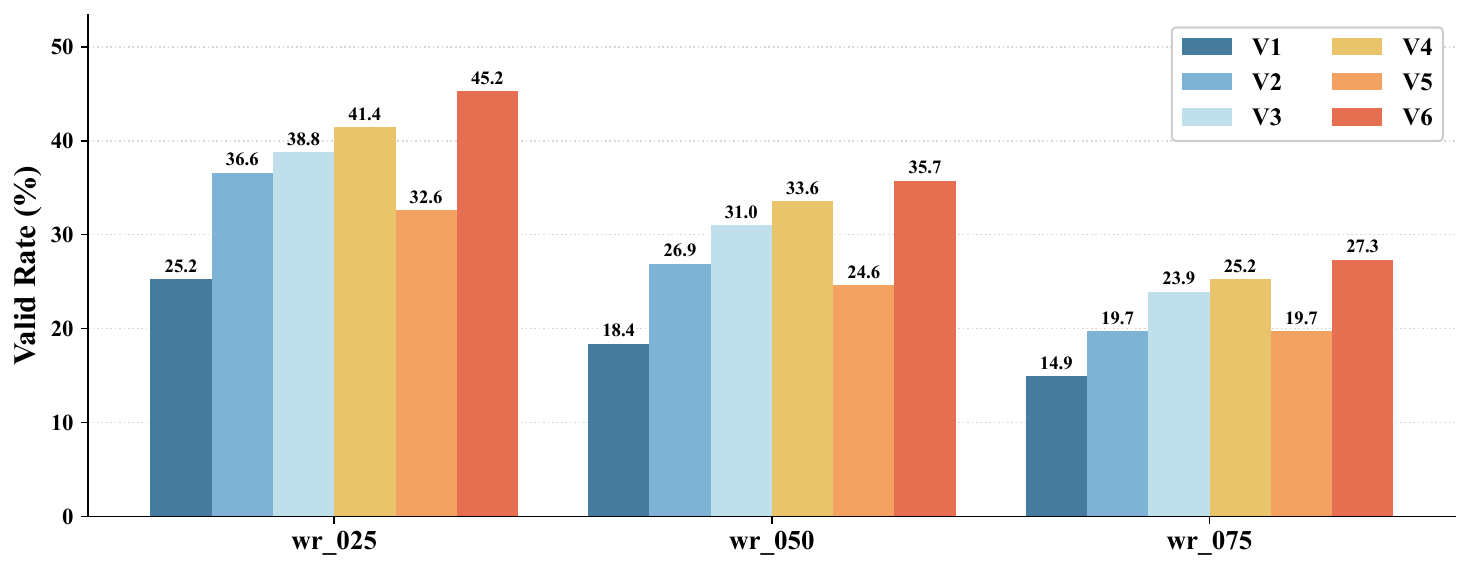}
\caption{Total valid rate of the six context settings on the three
evaluation benchmarks (\texttt{wr\_025}, \texttt{wr\_050},
\texttt{wr\_075}), aggregated across $3$-, $4$-, $5$-agent
instances and all map sizes. V6 achieves the highest valid rate on
every benchmark.}
\label{fig:valid_rate}
\end{figure*}

Figure~\ref{fig:valid_rate} shows the total valid rate of all six
context settings on the three evaluation benchmarks, aggregated
across $3$-, $4$-, $5$-agent instances and all map sizes. V6
achieves the highest valid rate on every benchmark.

\section{Training details}
\label{app:training_details}

We train with GRPO on top of Qwen3-4B (base). Each GRPO update
draws a batch of $128$ prompts; for every prompt we sample $n{=}8$
trajectories from the current actor with vLLM
(\texttt{tensor\_parallel}${=}2$,
\texttt{gpu\_memory\_utilization}${=}0.55$,
\texttt{max\_prompt\_length}${=}1536$,
\texttt{max\_response\_length}${=}5120$). The resulting $1024$
samples form one mini-batch, processed with per-GPU micro-batch
$4$, yielding $8$ steps of gradient accumulation across
$4{\times}$ H$100$ $80$\,GB GPUs.

We optimize the actor with AdamW at a constant learning rate of
$2{\times}10^{-6}$ and apply a low-variance KL penalty
($\beta{=}10^{-3}$) directly in the actor loss; no KL is added to
the reward and entropy regularization is disabled. GRPO requires
no critic. The reference policy is the policy at the start of
each round and is kept frozen with FSDP parameter offload; the
actor uses FSDP without offload but with gradient checkpointing.

Each training round performs three RL epochs over the round's
$4000$-sample training set, and the pipeline runs for three
rounds in total. The software stack is VERL v$0.6.1$ with PyTorch
$2.8$ / CUDA $12.4$ / vLLM $0.10.3$.

\section{Prompts used by the environment engineer}
\label{app:prompts}

This section reproduces the two prompts used at every training
round. The \emph{analysis prompt} is given the evaluation
breakdown and emits the next-round configuration as a fenced YAML
block. The \emph{summary prompt} condenses the just-completed
round into a short note that is then fed back as
\textbf{S}-module context.

\begin{tcolorbox}[
  enhanced, breakable,
  colback=gray!10, colframe=gray!10,
  coltitle=white, colbacktitle=gray!55,
  fonttitle=\bfseries\sffamily,
  title=Analysis prompt,
  arc=2pt, boxrule=0pt,
  left=8pt, right=8pt, top=6pt, bottom=6pt,
  before skip=10pt, after skip=10pt,
]
\begin{Verbatim}[fontsize=\footnotesize, breaklines=true, breakanywhere=true]
You are analyzing the training progress of an AI model that learns to solve Multi-Agent Path Finding (MAPF) problems.

## Task Description
MAPF involves finding collision-free paths for multiple agents on a grid map with obstacles (holes). Each agent must move from start to goal. At each time step, an agent can move UP, DOWN, LEFT, RIGHT, or WAIT. Two agents cannot occupy the same cell (vertex conflict) or swap positions (edge conflict).

The model is trained exclusively on 2-agent data but evaluated on 3/4/5-agent tasks to test generalization. The goal is for the model to learn fundamental MAPF skills (spatial reasoning, obstacle avoidance, conflict resolution) from 2-agent scenarios and generalize to more agents. Your config should only control 2-agent training data -- do NOT add 3/4/5-agent entries.
{training_details}
## Training Data Configuration
The total number of training samples is FIXED at 4000 across all map sizes. You control how those 4000 samples are distributed by setting per-map-size **ratios**.

The training data is generated from a YAML config that controls:
- **Map sizes** (e.g. 3x3, 5x5, 10x10): The grid dimensions for the MAPF problem.
- **ratio** per size: Share of the 4000 samples allocated to that map size (range 0-1). ...
- **hole_ratio** per size: Fraction of cells that are obstacles (range 0-0.5).
- **wait_ratio** per size: Fraction of samples that must contain WAIT actions for conflict resolution (range 0-1).

You can adjust any of these parameters to tailor the training curriculum. ...

## Training History
{history_section}

## Your Task
Analyze the evaluation results above. You are given:
1. **Overall failure breakdown**: failure counts across all map sizes (categories are NOT mutually exclusive).
2. **Per map-size summary**: Valid% and Optimal% for each map size, showing where the model excels vs. struggles.
3. **Per map-size failure breakdown**: failure type counts broken down by map size.

Failure type reference:
- **Conflict failures**: two agents collided (vertex or edge conflict).
- **Parse failures**: the model output could not be parsed into valid action sequences.
- **Illegal moves**: the model output an action that is not in the action space.
- **Goal failures**: one or more agents did not reach their goal position.
- **Hole collisions**: an agent moved onto an obstacle cell.
- **Out of bounds**: an agent moved outside the grid.

## Guidelines
- If training history is provided, pay close attention to how the failure distribution AND per-size performance changed after each config adjustment.
- Since training is continual, overall metrics will generally improve over rounds. Pay more attention to which map sizes or failure types are still lagging behind ...
- Prior config adjustments were made based on earlier training stages and may no longer be appropriate. Make your own judgment based on the current metrics ...
- You may also analyze the current training data distribution to identify what types of data are missing or underrepresented ...
- **IMPORTANT**: The model learns progressively from easy to hard -- ensure it has mastered simpler tasks before shifting focus to harder ones.
- **IMPORTANT**: Because the total budget is fixed at 4000 samples and all 8 sizes (3x3 to 10x10) are always included, increasing the ratio for one map size implicitly takes samples away from another. Think of the 8 ratios as a probability distribution that always sums to 1.0.

Based on your analysis (and the trend if history is available), recommend changes to the generation config for the next training round. You may change ratio, hole_ratio, and/or wait_ratio for any map sizes.

Output your recommendation as a fenced YAML block (```yaml ... ```). You do NOT need to output every field -- only the fields you want to change. ...

The YAML structure is:
```yaml
generation:
  2_agents:
    NxN: {{ratio: <float 0-1>, hole_ratio: <float 0-0.5>, wait_ratio: <float 0-1>}}
```
\end{Verbatim}
\end{tcolorbox}

\begin{tcolorbox}[
  enhanced, breakable,
  colback=gray!10, colframe=gray!10,
  coltitle=white, colbacktitle=gray!55,
  fonttitle=\bfseries\sffamily,
  title=Summary prompt,
  arc=2pt, boxrule=0pt,
  left=8pt, right=8pt, top=6pt, bottom=6pt,
  before skip=10pt, after skip=10pt,
]
\begin{Verbatim}[fontsize=\footnotesize, breaklines=true, breakanywhere=true]
You are reviewing the training progress of an AI model learning to solve Multi-Agent Path Finding (MAPF) problems.

## Background
MAPF involves finding collision-free paths for multiple agents on a grid map. The training data is generated from a YAML config controlling map sizes, sample counts, hole_ratio (obstacle density), and wait_ratio (fraction of conflict-resolution samples).

## Training History
{history_section}

## New Config Chosen
{new_config_summary}

## Your Task
Based on the evaluation results and the new config above, write a concise summary (3-5 sentences) explaining:
1. What are the key observations from the current evaluation results?
2. Why was this config chosen -- what weaknesses is it trying to address?
3. What should be monitored in the next round to judge if this adjustment was effective?

Be specific about map sizes, failure types, and parameter changes. Do NOT output any YAML or config -- only your reasoning summary.
\end{Verbatim}
\end{tcolorbox}


\section{Limitations}
Our study has several limitations that suggest directions for future work. First, while MAPF-FrozenLake provides fine-grained control over training distributions, it represents a single, self-contained task family. The redesign strategies learned by the environment engineer may not directly transfer to domains with qualitatively different failure modes or evaluation signals. Second, our experiments focus on a specific RL training pipeline; the interaction between environment redesign and other training paradigms, such as online imitation learning or reward-free exploration, remains unexplored. Finally, our current framework restricts the environment engineer to a fixed generator architecture. Extending the approach to allow structural modifications of the generator itself, such as introducing new environment mechanics, raises additional challenges that we do not address here.

\end{document}